\documentclass{evolang11}
\usepackage{color}
\usepackage{graphicx}
\usepackage{url}
\usepackage{etoolbox} 

\usepackage{fancyhdr}

\begin{document}

\newtoggle{anonymous}
\togglefalse{anonymous}

\newtoggle{arxiv}
\toggletrue{arxiv}

\iftoggle{arxiv}{
  \thispagestyle{plain}
  \pagestyle{plain} 
  \addtolength\footskip{3em}
  \cfoot{\phantom{x}\\\thepage}
}

\title{Kauffman's adjacent possible in word order evolution}

\iftoggle{anonymous}{
   \author{ANONYMOUS AUTHOR}
   \address{University Department, University Name \\ City, Country\\email@university}
}{
   \author{Ramon Ferrer-i-Cancho}
   \address{Departament de Ci\`encies de la Computaci\'o, Universitat Polit\`ecnica de Catalunya \\ Barcelona, Catalonia (Spain) \\rferrericancho@cs.upc.edu}  
}

\maketitle

\abstracts{Word order evolution has been hypothesized to be constrained by a word order permutation ring: transitions involving orders that are closer in the permutation ring are more likely. The hypothesis can be seen as a particular case of Kauffman's adjacent possible in word order evolution.   
Here we consider the problem of the association of the six possible orders of S, V and O to yield a couple of primary alternating orders as a window to word order evolution. We evaluate the suitability of various competing hypotheses to predict one member of the couple from the other with the help of information theoretic model selection. Our ensemble of models includes a six-way model that is based on the word order permutation ring (Kauffman's adjacent possible) and another model based on the dual two-way of standard typology, that reduces word order to basic orders preferences (e.g., a preference for SV over VS and another for SO over OS). Our analysis indicates that the permutation ring yields the best model when favoring parsimony strongly, providing support for Kauffman's general view and a six-way typology. }

\section{Introduction}

\label{introduction_section}

It is widely agreed that cultural evolution is more important than cognitive factors in governing word order change \cite{Dryer2011b,Dunn2011a,Gell-Mann2011a}. For instance, Dryer wrote that {\em ``Dunn et al. are right that cultural evolution is more important than cognitive factors in governing word order change''} while Gell-Mann \& Ruhlen wrote that {\em ``we know of no evidence that SOV, SVO, or any other word order confers any selective advantage in evolution''.}

However, word order evolution exhibits some trends: change has been for the most part from SOV to SVO and, beyond that, from SVO to VSO/VOS with a subsequent reversion to SVO occurring occasionally \cite{Gell-Mann2011a}. 
Are those trends not reflecting any cognitive bias? Cultural evolution and cognitive biases could be both at play.
 
Let us consider the case of the principle of dependency length minimization, which predicts that 
the optimal placement of the verb is at the center, e.g. SVO or OVS \cite{Ferrer2008e,Ferrer2013e}. 
A proponent of that cognitive bias in word order evolution has to solve various puzzles, for instance:
\begin{enumerate}
\item
Why the transition from SOV to OVS (one of the two orders where the verb is at the center), is rare \cite{Gell-Mann2011a}. If that principle is strong enough to pull SOV towards SVO it should be able to do the same job for SOV towards OVS.  
\item
If the principle is so strong to drive word order change towards SVO, why some languages have left SVO towards VSO/VOS \cite{Gell-Mann2011a}? 
\end{enumerate}
The first puzzle has been solved with the help of the concept of word order permutation ring, a circular graph where two word orders are connected if one yields the other by a swap of adjacent constituents \cite{Ferrer2008e,Ferrer2013e}. That ring allows one to define a distance between word orders and to hypothesize that the likelihood of a transition is a decreasing function of the distance between the orders, which is justified as a prediction of a generalized principle of Euclidean distance minimization \cite{Ferrer2013e}. According to this hypothesis, SOV $\rightarrow$ SVO is more likely than SOV $\rightarrow$ OVS \cite{Ferrer2013e}. See the supplementary materials for a further justification and some clarifications on the ring hypothesis. 

The second puzzle can be solved combining two facts. The first fact is that word order is a multiconstraint satisfaction problem and VSO/VOS is not optimal from the point of view of dependency length minimization but optimal from the perspective of maximizing the predictability of the contents of the nominal phrases (S and O) \cite{Ferrer2014a}. The second fact is a predisposition of SVO order to VSO/VOS. From the perspective of dependency length minimization, SVO languages could put adjectives and other dependents of the nominal heads of S and O either before and after those heads. However, postnominal placements prevent regression to SOV because SOV prefers prenominal placements. To escape from SOV, it is convenient for a SVO language to place dependents of nominal head after the verb which, interestingly, is the preferred placement for VSO/VOS \cite{Ferrer2013e}. Thus, escaping from SOV towards SVO and stabilizing the change can preadapt a SVO order to become VSO/VOS. Again, the principle of dependency length minimization is a key factor. 

The subtle predictions of dependency length minimization (and other word order principles) reviewed above suggest that 
the conclusion that cultural evolution is more important than cognitive factors comes from a hidden assumption: that cognitive factors apply absolutely, regardless of the current state of a language (e.g., its current dominant word order) and its nearest states (e.g., the nearest dominant word orders). 
In absolute terms (not conditioning on the current state), dependency length minimization predicts SVO/OVS. In local or relative terms (conditioning on the current state), dependency length minimization predicts that SVO is more likely than OVS when the current state is SOV, in full agreement with {\em ``the current state of a linguistic system shaping and constraining future states"} \cite{Dunn2011a}.
These converging relativistic views can be seen as examples of Kauffman's adjacent possible: the current configuration of a system can determine the next states, overriding prior unconditional biases \cite{Kauffman2000a}. 

In this article we consider the case of languages that do not exhibit a single dominant ordering of the triple but rather two primary  alternating word orders (Table \ref{primary_alternating_orders_table}), e.g., SOV/SVO in case of German \cite{wals-81}.  
Those languages are a subset of the languages lacking a dominant word order: there are 189 languages lacking a dominant word order but only 67 exhibit such a couple of primary alternating orders \cite{wals-81}.
Here we investigate why an order associates with another to form a couple of primary alternating word orders as a mirror of constraints on word order variation and evolution. In particular, we would like to know if those associations are constrained by the distance between the partners in the word order permutation ring, being closer associations more likely, in agreement with Kauffman's adjacent possible. If that was the case, that would mean that the restrictions imposed by the ring are pervasive and then fundamental for the development of a parsimonious theory of language and word order in particular. The relationship between the problem of primary alternating word orders and the evolution of language becomes evident under the assumption that word order change exploits constraints on word order variation. If word order evolution is actually constrained by the word order permutation ring diachronically as we have explained above, we should also observe the manifestation of this ring in word order variation synchronically.

We consider various hypotheses for the choice of an order $y$ as a partner given that the other partner is order $x$. We begin with a general and intuitive presentation (full mathematical details are given in Section \ref{models_section}).   
\begin{itemize}
\item
Model 0: the null hypothesis of obtaining $y$ by rolling a die (one side for every ordering of the triple) till $y \neq x$. 
\item
Model 1: a model where the probability of choosing $y$ depends on its distance to $x$ in the permutation ring mentioned above.
\item
Models 2-4: a random choice proportional to {\em a priori} probabilities of each of the six orders.
\end{itemize}
Model 1 and all the arguments on word order transitions reviewed above are examples of a six-way approach (the six possible orderings of $S$, $V$ and $O$) that is in stark contrast to the dual two-way view of linguistic typology, where it is believed that linguistic  properties are a direct consequence of preferences over pairs of basic orders \cite{Cysouw2008a}. For instance, according to the latter, the frequency of each of the 6 possible orderings of $S$, $V$ and $O$ is argued to originate from \cite{Cysouw2008a}
\begin{itemize}
\item 
A preference for SV over VS.
\item
A preference for SO over OS.
\item
No preference for OV over VO.  
\end{itemize}
In Model 2, the {\em a priori} probabilities reflect the dual two-way view. The parameters of the model define the probability for every basic word order. 
Model 3 is a six-way model where the {\em a priori} probabilities of each triple are derived from the probability of association of the triples in the real dataset. 
In contrast, Model 4 is another six-way model where {\em a priori} probabilities are defined by the relative frequency of the six orders as dominant orders, a factor considered to be relevant when analyzing primary alternating words orders \cite{wals-81}.
 
It is important to notice that all the models except Model 1 make absolute predictions (in the sense that the knowledge about $x$ only imposes that $y \neq x$ since the primary alternating orders must be different by definition). Model 1 makes a truly relative prediction because it takes into account the distance between $x$ and $y$ in the word order permutation ring. Furthermore, Model 1 is the only model with a clear cost-cutting hypothesis behind. Our analysis based on information theoretic model selection and further concerns for parsimony will show that Model 1 is the best, calling for a revision of the view that cognitive factors are secondary in word order change \cite{Dryer2011b,Dunn2011a,Gell-Mann2011a} and providing new support for the six-way approach in the debate between six-way and two-way typology \cite{Newmeyer2005a,Dryer2013a}. 
 
\begin{table}[ht]
   \tablecaption{Primary alternating orders of $S$, $V$ and $O$ according to the World Atlas of Language Structures {\protect \cite{wals-81}}. $m(x,y)$ is the number of languages where orders $x$ and $y$ are primary alternating orders.
$d(x,y)$ is the distance between $x$ and $y$ in the permutation ring {\protect \cite{Ferrer2013e}}. }
{\footnotesize
\begin{tabular}{@{}cccc@{}}
$x$ & $y$ & $m(x,y)$ & $d(x,y)$ \\
\hline
SOV & SVO & 29 & 1 \\
VSO & VOS & 14 & 1 \\
SVO & VSO & 13 & 1 \\
SVO & VOS & 8 & 2 \\
SOV & OVS & 3 & 2 \\
\end{tabular}
\label{primary_alternating_orders_table}
}
\end{table}

\section{The models}

\label{models_section}

${\cal P}_1$ is defined as the set of all the possible orderings of $S$, $V$ and $O$, i.e. 
\begin{equation}
{\cal P}_1 =  \{SOV, SVO, VSO, VOS, OVS, OSV \}.
\end{equation}
${\cal P}_2$ is defined as the set of all the different unordered pairs of elements of ${\cal P}_1$. 
The cardinality of ${\cal P}_1$, $|{\cal P}_1|$, is $6$, and 
$|{\cal P}_2| = {6 \choose 2} = 15$. 

Suppose that $p(y|x)$ is the probability that $y$ is chosen as a partner knowing that $x$ is already one of the partners in the couple of primary alternating orders of a language. $p(y|x)$ is defined over ${\cal P}_2$ ($p(y|x) = 0$ if $(x,y) \notin {\cal P}_2$). We use $\pi$ to refer to the parameters of a model (if any). 
We consider the following models for $p(y|x)$: 
\begin{itemize}
\item
{\em Model 0}. \\ A null model where $p(y|x) = 1/5$. This model has 0 parameters.
\item
{\em Model 1}. \\ Suppose that $k(d)$ indicates the number of neighbors of an order at distance $d$ in the permutation ring. It is easy to see that $k(d) = 2$ if $d \in \{1, 2\}$ and $k(d) = 1$ if $d = 3$. Suppose that $d(x,y)$ is the distance between orders $x$ and $y$ in the permutation ring (Table \ref{primary_alternating_orders_table}). This model defines $p(y|x)$ as
\begin{equation}
p(y|x) = \frac{p(d(x,y))}{k(d(x,y))},
\label{adjacent_possible_model_equation}
\end{equation}
where $p(d)$ is the probability that an association involves orders at distance $d$ in the permutation ring. 
The $1/k(d(x,y))$ factor in Eq. \ref{adjacent_possible_model_equation} indicates that Model 1 chooses the partner at distance $d(x,y)$ uniformly at random.  
Model 1 defines $p(d)$ with two parameters $\pi(1)$ and $\pi(2)$ such that 
$p(1) = \pi(1)$, $p(2) = \pi(2)$ and $p(3) = 1 - \pi(1) - \pi(2)$.   
\item
{\em Models 2-4}. These three models assume
\begin{equation}
p(y|x) \propto q(y),
\label{conditional_probability_equation}
\end{equation}
where $q(y)$ is some {\em a priori} probability of $y$. Each of the three models defines $q(y)$ in a different way (details of each below).  
 
\end{itemize}
In standard typology, the choice of a possible ordering of $S$, $V$ and $O$ is believed to originate from local word order preferences that can defined by $p(\alpha\beta)$, the probability of placing $\alpha$ before $\beta$ (not necessarily consecutively), with $p(\beta \alpha) = 1 - p(\alpha \beta)$. Model 2 has three parameters, i.e. $\pi(SV)$, $\pi(SO)$ and $\pi(OV)$, which define $p(SV)$, $p(SO)$ and $p(OV)$, respectively. In Model 2, $q(y) = c r(y)$,
where $c$ is a normalization constant and
\begin{eqnarray}
r(SOV) & = & c \cdot p(SV)p(SO)p(OV) \nonumber \\
r(SVO) & = & c \cdot p(SV)p(SO)(1 - p(OV)) \nonumber \\
r(VSO) & = & c \cdot (1 - p(SV))p(SO)(1 - p(OV)) \nonumber \\
r(VOS) & = & c \cdot (1 - p(SV))(1 - p(SO))(1 - p(OV)) \nonumber \\
r(OVS) & = & c \cdot (1 - p(SV))(1 - p(SO))p(OV) \nonumber \\
r(OSV) & = & c \cdot p(SV)(1 - p(SO))p(OV).
\end{eqnarray}
Model 3 defines $q(y)$ with five parameters, e.g., $\pi(y)$ for $y \in {\cal P}_1 \setminus \{ OSV \}$ since
\begin{equation}
q(SOV) = 1 - \sum_{y \in {\cal P}_1 \setminus \{ OSV \}} \pi(y).
\end{equation}    
Model 4 is a particular case of Model 3 where $q(y)$ is the relative frequency of languages where $y$ is dominant \cite{wals-81}. Model 4 has no parameters. Model 0 is provided as a control for Models 1-4. Models 1-4 should be better than Model 0, the null hypothesis. See the supplementary materials for further mathematical details about the models.

\section{Evaluation}

$m(x,y)$ is the number of languages where $x$ and $y$ are primary alternating orders, with $m(x,y) = m(y,x)$ (Table \ref{primary_alternating_orders_table}). 
The total number of languages is defined as
\begin{equation}
m = \sum_{(x,y) \in {\cal P}_2} m(x,y) = \frac{1}{2} \sum_{(x,y) \in {\cal P}_1 \times {\cal P}_1, x \neq y} m(x,y),
\end{equation}
where ${\cal P}_1 \times {\cal P}_1$ is the Cartesian product of ${\cal P}_1$ with itself. 
The actual sample size is $n = 2m$ since every language in the sample has two orders in partnership and our models predict each.
 
The log-likelihood of every partner in the primary alternating orders of a set of languages is defined as  
\begin{equation}
{\cal L} = \sum_{(x,y) \in {\cal P}_1 \times {\cal P}_1, x \neq y} m(x,y) \log p(y|x). 
\end{equation}
We evaluate the quality of the fit of a model with $k$ parameters over a sample of size $n$ with two scores. One is the corrected Akaike information criterion \cite{Burnham2002a}, defined as
\begin{equation}
AIC_c = -2 {\cal L} + 2\frac{kn}{n-k-1}. \label{AIC_c_equation}
\end{equation}
The other is the Bayesian information criterion \cite{Wagenmakers2004a}, defined as 
\begin{equation}
BIC = -2 {\cal L} + k \log n. \label{BIC_equation}
\end{equation}

The smaller the score, the better the model \cite{Burnham2002a}. We use two criteria and not simply one to show overtly the assumptions of our arguments and reflect about them. AIC$_c$ and BIC offer complementary perspectives \cite{Wagenmakers2004a}: 
\begin{itemize}
\item
While BIC assumes that the true generation model is in the set of candidate models, AIC$_c$ does not assume that any of the candidate models is necessarily true. 
\item
When $n \geq \lceil e^2 \rceil = 8$, BIC introduces a stronger penalization for the number of parameters than AIC$_c$ (see supplementary materials for a detailed argument). Here $n = 2m = 134$ since $m = 67$ according to Table \ref{primary_alternating_orders_table} and thus BIC is going to penalize more than AIC$_c$. This is of special relevance here because in case that BIC and AIC$_c$ do not agree on the best model, we will assume that reality is relatively low dimensional or that the best model has to be inserted in a compact general theory of language. As a result of these assumptions, we will take the best model according to BIC as the final best, as this criterion favors the model with the smallest number of parameters in our dataset. 
\end{itemize}
 
The best parameters of every model are estimated replacing every probability parameter by the corresponding proportion in the dataset in Table \ref{primary_alternating_orders_table} (see supplementary materials for further details about parameter estimation).

\section{Results}

Table \ref{evaluation_of_models_table} summarizes the results of the evaluation of the different models.
Model 3 is the best model according to AIC$_c$ whereas Model 1 is the best model according to BIC. These two models are in a kind of tie because the difference of AIC$_c$ between Model 1 and 3, as well as that of BIC, are small. However, we favor the winner according to BIC for two reasons. On a local basis, Model 1 has less parameters than Model 3. On a general basis, Model 1 leads to a lighter general theory of language: with Model 1, we can explain both the diachronic patterns reviewed in Section \ref{introduction_section} but also synchronic patterns such as the primary alternating orders. If we chose Model 3, the general theory becomes fatter because Model 3 has more parameters than model 1 and also because additional arguments to explain the origins of these diachronic patters are required.      
Interestingly, Model 2, the model inspired by standard typology, is worse than Model 1 and 3 according to both AIC$_c$ and BIC. Model 4 is even worse than Model 0, the null hypothesis.

The fact that Model 1 is the final best model does not imply that the word order permutation hypothesis is valid because this hypothesis means $p(1) > p(2) > p(3)$. Actual support for the word order permutation is provided by the maximum likelihood estimates, which are  $\pi(1) = 0.84$, $\pi(2) = 0.16$. The latter finding suggests that word order associations for dominant word orders are constrained in the same way as word order transitions during evolution. As for Model 2, the maximum likelihood parameters are $\pi(SV) = 0.61$, $\pi(SO) = 0.81$ and $\pi(OV) = 0.26$.


\begin{table}[ht]
  \tablecaption{Summary of the evaluation of the models. ${\cal L}$ is the log-likelihood, $k$ is the number of parameters of the model, AIC$_c$ is the corrected AIC and BIC is the Bayesian information criterion. }
{\footnotesize
\begin{tabular}{@{}ccccc@{}}
Model & ${\cal L}$ & $k$ & AIC$_c$ & BIC \\
\hline
0 & -215.7 & 0 & 431.3 & 431.3 \\
1 & -152.7 & 2 & 309.5 & 315.2 \\
2 & -161.5 & 3 & 329.3 & 337.8 \\
3 & -147.3 & 5 & 305.2 & 319.2 \\
4 & -285.3 & 0 & 570.6 & 570.6 \\
\end{tabular}
\label{evaluation_of_models_table}
}
\end{table}

\section{Discussion}

We have shown that Model 1 is the best model when favoring parsimony strongly on a local basis (using BIC) and also on a general basis (favoring a light general theory of language). Model 1 has beaten all the models that assume that word order biases apply independently from the current state of the system. This is not very surprising since 
those models suffer from a strong bias for SVO or SOV that has difficulties to deal with the 14 languages where VSO and VOS are the primary dominant orders (Table \ref{evaluation_of_models_table}).   
Some parameter reductions of Models 1 and 2 are discussed in the supplementary materials. That of model 1 is specially relevant because it could increase the difference in BIC between Model 1 and 3, which is currently rather small (Table \ref{evaluation_of_models_table}).   

We have presented an analysis that shows the potential of information theoretic model selection \cite{Burnham2002a} to evaluate competing hypotheses about word order variation. The failure of Model 2 challenges the strong belief in the predictive power of a dual two-way approach of standard typology \cite{Cysouw2008a,Dryer2013a}. Six-way approaches \cite{Newmeyer2005a} should be reconsidered. 
The success of Model 1 when favoring parsimony strongly suggests that the view that cognitive cost minimization is secondary to word order change \cite{Dryer2011b,Dunn2011a,Gell-Mann2011a} should be revised.

All these considerations suggest that Kauffman's notion of adjacent possible \cite{Kauffman2000a} is for crucial for progress in typology and word order evolution research. Although our work has an important limitation, i.e. it assumes that languages are independent \cite{Cysouw2008a}, we hope to have built ``adjacent possibles'' to apply, hopefully in a near future, the powerful phylogenetic methods that have already been used to investigate word order evolution with the traditional dual two-way approach \cite{Dunn2011a}.  

\iftoggle{anonymous}{

}{

  \section*{Acknowledgments}

  The feedback of three anonymous reviewers has been crucial. 
  We are also grateful to B. Bickel, W. Croft, E. Santacreu-Vasut and S. Strogatz for helpful discussions. 
  This research has been funded by the grants 2014SGR 890 (MACDA) from AGAUR (Generalitat de Catalunya) and the grant 
  TIN2014-57226-P from MINECO (Ministerio de Economia y Competitividad).
}

\bibliographystyle{apacite} 
\bibliography{../biblio/complex,../biblio/rferrericancho,../biblio/ling,../biblio/cs,../biblio/cl,../biblio/maths}

\iftoggle{arxiv}{

\pagebreak

\appendix

\noindent {\bf \Large SUPPLEMENTARY ONLINE INFORMATION}

   \iftoggle{arxiv}{
    \section*{The word order permutation ring hypothesis}
}{
    \section{The word order permutation ring hypothesis}
}

In the word order permutation ring \cite{Ferrer2008e,Ferrer2013e}, two orders are connected if one yields the other by swapping a couple of adjacent constituents (Fig. \ref{permutation_ring_figure}).
The word order permutation ring allows one to define a distance between a couple of word orders: the minimum number of edges that are needed to reach one order from the other, which is equivalent to the number of swaps of adjacent constituents that are needed to transform one order into the other. For instance, SOV and SVO are at distance one while SOV and OVS are at distance 2. 
This distance between orders (measured in edges in the ring) is used to hypothesize that the probability of a transition decreases as distance increases \cite{Ferrer2013e}. The hypothesis does not claim that word order change proceeds by swapping adjacent constituents exclusively. For instance, when the ring indicates that SOV and OVS are at distance two it does not mean that the actual change must be in two steps (with one swap of adjacent of constituents each): SOV to OSV and OSV to OVS. The hypothesis is neutral concerning whether changes at distances 2 or 3 are made via swaps of adjacent constituents or not. It only assumes one step for changes at distance one. Interestingly, it is well known that the transition from SOV to OVS takes place in one step \cite{Gell-Mann2011a}, which is fully compatible with the permutation ring hypothesis.  

\begin{figure}
\begin{center}
\includegraphics[scale = 0.6]{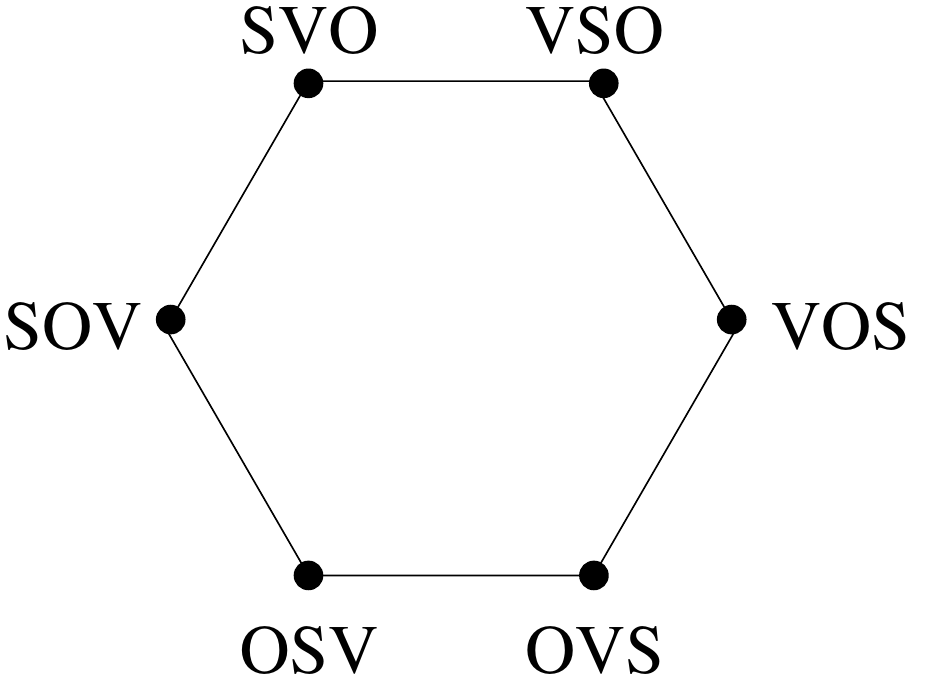}
\end{center}
\caption{\label{permutation_ring_figure} The word order permutation ring \protect \cite{Ferrer2013e}. }
\end{figure}

The permutation ring has been regarded as a prediction of the principle of Euclidean distance minimization \cite{Ferrer2013e} in the sense that a swap of adjacent constituents is a rearrangement where the elements moved and their new positions are as close as possible. A further connection with that general principle of sequential processing becomes more evident when considering $\Delta(\sigma,\sigma')$, a new measure of the distance between $\sigma$ and $\sigma'$, two orderings of the same triple such that $\sigma = xyz$ and $\sigma'$ is a permutation of $\sigma$. Suppose that $\delta_x$ is the distance between the two placements of $x$ in $\sigma$ and $\sigma'$. For instance, if $\sigma = SOV$ and $\sigma' = SVO$ then $\delta_S = 0$, $\delta_O = \delta_V = 1$; if $\sigma = SOV$ and $\sigma' = VOS$ then $\delta_S = , \delta_V = 2$ and $\delta_O = 0$. $\Delta$ is defined as the total displacement of elements between $\sigma$ and $\sigma'$, i.e.
\begin{equation}
\Delta = \delta_x + \delta_y + \delta_z.
\end{equation}
The general principle of Euclidean distance minimization \cite{Ferrer2004b} predicts that $\Delta$, the Euclidean distance between the placements of each constituent, is also minimized. Table \ref{distances_table} compares $\Delta$ against $d$, the distance in the word order permutation ring. According to Table \ref{distances_table}, both distances are positively correlated but not perfectly: 
their Kendall $\tau$ correlation is 0.92 with a sample size of 6 and a p-value of 0.03. The only obstacle for a perfect correlation is that $\Delta = 4$ when $d = 2$ or $d = 3$.  
Therefore, although $d$ measures a distance in a network space and the principle of dependency length minimization measures a distance in a 1-dimensional space, $d$ allows, approximately, an interpretation as a distance in a 1-dimensional space too.  
The importance of distance for a theory of language synchrony and diachrony is reminiscent of Zipf's metaphor of the distance between a craftsman and its tools, which also has to be minimized \cite{Zipf1949a}.  

\begin{table}
    \tablecaption{Summary of the distance between an order $\sigma = xyz$ and an order $\sigma'$, a permutation of $\sigma'$. Two measures are considered: $d$, the distance in the word order permutation ring and $\Delta$, the total displacement. }
{\footnotesize 
\begin{tabular}{@{}cccccc@{}}
$\sigma'$ & $\delta_x$ & $\delta_y$ & $\delta_z$ & $\Delta(\sigma,\sigma')$ & $d(\sigma,\sigma')$ \\  
\hline  
xyz       & 0 & 0 & 0 & 0 & 0 \\
xzy       & 0 & 1 & 1 & 2 & 1 \\
yxz       & 1 & 1 & 0 & 2 & 1 \\
zxy       & 1 & 1 & 2 & 4 & 2 \\
yzx       & 2 & 1 & 1 & 4 & 2 \\
zyx       & 2 & 0 & 2 & 4 & 3 \\
\end{tabular}
\label{distances_table}
}
\end{table}

 
\iftoggle{arxiv}{
    \section*{Further details about Model 1}
}{
    \section{Further details about Model 1}
}

It is easy to check that Model 1 defines a true probability for $p(y|x)$, i.e. 
\begin{equation}
\sum_{y} p(y|x) = 1. 
\end{equation}
Applying the definition of $p(y|x)$ of Model 1, we obtain
\begin{eqnarray}
\sum_{y} p(y|x) & = & \sum_{y} \frac{p(d(x,y))}{k(d(x,y))} \nonumber \\
                & = & k(1) \frac{\pi(1)}{k(1)} + k(2) \frac{\pi(2)}{k(2)} + k(3) \frac{p(3)}{k(3)} \nonumber \\
                & = & \pi(1) + \pi(2) + p(3) \nonumber \\
                & = & 1,  
\end{eqnarray} 
as we wanted to prove.   

It is easy to show that Model 0 is a particular case of Model 1 when $\pi(1) = \pi(2) = 2/5$ and then $p(3) = 1 - \pi(1) - \pi(2) = 1/5$.  
Suppose that Model 0 and Model 1 are the same, namely 
\begin{equation}
p(y|x) = \frac{1}{5} = \frac{p(d(x,y))}{k(d(x,y))}.
\label{adjacent_possible_model_supplementary_equation}
\end{equation}
If $d(x,y) = 1$ or $d(x,y) = 2$ we have $k(d(x,y)) = 1$, which transforms Eq. \ref{adjacent_possible_model_supplementary_equation} into
\begin{equation}
\frac{1}{5} = \frac{\pi(1)}{2} = \frac{\pi(2)}{2}.
\end{equation}
Thus, $\pi(1) = \pi(2) = \frac{2}{5}$. If $d(x,y) = 3$ we have $k(d(x,y)) = 1$, which transforms Eq. \ref{adjacent_possible_model_supplementary_equation} into
\begin{equation}
\frac{1}{5} = p(3)
\end{equation}
as expected.

Model 1 could be improved reducing the number of parameters to 1. This could be achieved modeling $\pi(d)$ for $1 \leq d \leq 3$ with a distribution with a single parameter, e.g., a right truncated exponential distribution, defined as 
\begin{equation}
\pi(d) = c e^{-ad},
\end{equation}
where $c$ is a normalization constant, $a$ is a parameter and $1 \leq d \leq 3$.  
Such a reduction of parameters could decrease the values of AIC$_c$ and BIC for Model 1 (recall definition of these scores in the main article) and, interestingly, it could increase the distance between that model and its closest competitor, Model 3, which is rather small at present.

\iftoggle{arxiv}{
    \section*{Further details about Models 2-4}
}{
    \section{Further details about Models 2-4}
}

Models 2-4 assume
\begin{equation}
p(y|x) \propto q(y).
\label{conditional_probability_supplementary_equation}
\end{equation}
Combining this assumption with the fact that 
\begin{equation}
\sum_{y} p(y| x) = 1 
\end{equation}
one obtains that 
\begin{equation}
p(y|x) = \frac{q(y)}{1-q(x)},
\end{equation}
where $1 - q(x)$ is the probability that $y \neq x$.

\iftoggle{arxiv}{
    \section*{Further details about Model 2}
}{
    \section{Further details about Model 2}
}

The fact that 
\begin{equation}
\sum_{x \in {\cal P}_1} q(x) = 1
\label{normalization_1_equation}
\end{equation}
by definition and  
\begin{equation}
\sum_{x \in {\cal P}_1} r(x) = p(SV)[p(OV) + p(SO) - 1] + 1 - p(SO)p(OV)
\end{equation}
gives
\begin{equation}
c = \frac{1}{p(SV)[p(OV) + p(SO) - 1] + 1 - p(SO)p(OV)}.
\end{equation}

We have translated the standard view of linguistic typology to a model with three parameters. As parsimony is a desirable property of a model, it is tempting to reduce the number of parameters to two: i.e. $\pi(SV)$ and $\pi(SO)$, inspired by the fact that there seems to be no preference for $OV$ over $VO$ \cite{Cysouw2008a}. Accordingly, we could simplify Model 2 imposing $p(OV) = p(VO) = 1/2$. The problem of this simplification is that the gain by the reduction of parameters is lost by a dramatic reduction of ${\cal L}$ and then the values of AIC$_c$ and BIC turn out to be greater than the ones obtained with the original Model 2 (${\cal L}$ = -175.9, AIC$_c$ = 356.0 and BIC = 361.7).

We could consider also another track to save one parameter: imposing $c = 1$ and inferring the third parameter $\pi(OV)$
from $p(SV)$ and $p(SO)$ via the normalization condition 
\begin{equation}
\sum_{x \in {\cal P}_2} p(x) = 1,
\label{normalization_2_equation}
\end{equation}
which gives
\begin{align}
\sum_{x \in {\cal P}_2} r(x) & = c^{-1} \nonumber \\ 
                             & = p(SV)[p(OV) + p(SO) - 1] + 1 - p(SO)p(OV) \nonumber \\
                             & = 1
\end{align}
and finally
\begin{equation}
p(OV) = \frac{p(SV)(1-p(SO))}{p(SV)-p(SO)}.
\label{probability_of_SO_equation}
\end{equation}
Applying the assumption that $p(\alpha\beta)$ is a probability we will unveil heavy constraints on $p(SV)$ and $p(SO)$. First, notice that the condition $p(OV) \leq 1$ implies $p(SV) \neq p(SO)$ because the numerator Eq. \ref{probability_of_SO_equation} is bounded by definition of $p(\alpha\beta)$ and the denominator needs $p(SV) \neq p(SO)$ to warrant $p(OV) \leq 1$. Assuming $p(SV) \neq p(SO)$, further constraints appear:
\begin{itemize}
\item  
The condition $p(OV) \leq 1$ also implies $p(SO)=1$. Notice that 
$0 \leq p(OV)$ is equivalent to $p(SO) \geq 1$ thanks to Eq. \ref{probability_of_SO_equation}. The fact that $p(SO)$ is a probability ($p(SO) \leq 1$) yields $p(SO)=1$. 
\item
The condition $0 \leq p(OV)$ implies $p(SV) > p(SO)$, a hard constraint on the parameter space. Notice that the numerator of the r.h.s. of Eq. \ref{probability_of_SO_equation} is positive by definition of $p(\alpha\beta)$ and then the denominator needs to be positive to satisfy $0 \leq p(OV)$. The assumption $p(SV) \neq p(SO)$ above gives $p(SV) > p(SO)$.  
\end{itemize}
Combining these results, the most serious caveat emerges: $p(SV) > 1$, i.e. $p(SV)$ is not a probability. 
Due to all these drawbacks, we have discarded a version of the model with just two parameters although the standard view (no preference for OV over VO \cite{Cysouw2008a}) suggests that two would suffice. 

\iftoggle{arxiv}{
    \section*{Information theoretic model selection}
}{
    \section{Information theoretic model selection}
}

The likelihood of every partner in the primary alternating orders of a set of languages is defined as
\begin{equation}
L = \prod_{(x,y) \in {\cal P}_1 \times {\cal P}_1, x \neq y} p(y|x)^{m(x,y)} 
\end{equation}
and then the log-likelihood is  
\begin{equation}
{\cal L} = \log L = \sum_{(x,y) \in {\cal P}_1 \times {\cal P}_1, x \neq y} m(x,y) \log p(y|x). 
\end{equation}

\begin{equation}
AIC_c = -2 {\cal L} + 2\frac{kn}{n-k-1}
\end{equation}
and 
\begin{equation}
BIC = -2 {\cal L} + k \log n
\end{equation} 
differ only in the penalty for lack of parsimony: while AIC$_c$ penalizes with 
\begin{equation} 
\frac{2k}{n - k - 1}, 
\end{equation}
BIC penalizes with 
\begin{equation}
k \log n.
\end{equation}
Thus, the penalty of AIC$_c$ is smaller than that of BIC if and only if 
\begin{equation}
\frac{2}{n - k - 1} < \log n.
\label{penalty_equation}
\end{equation} 
When $n \neq k + 1$ it is easy to see that   
\begin{equation}
\frac{2}{n - k - 1} \leq 2,
\end{equation}   
which allows one to conclude that Eq. \ref{penalty_equation} holds at least when 
\begin{equation}
n \geq \lceil e^2 \rceil = 8, 
\label{sample_size_equation}
\end{equation}
confirming a previous statement \cite{Wagenmakers2004a}.
Since the sample size of our dataset satisfies Eq. \ref{sample_size_equation}, we conclude that our dataset falls within the domain where BIC penalizes stronger for lack of parsimony than AIC$_c$.  

\iftoggle{arxiv}{
    \section*{Parameter estimation}
}{
    \section{Parameter estimation}
}

The best parameters of every model were estimated replacing every probability parameter by the corresponding proportion in the dataset. For instance, in Model 1, $\pi(1)$ and $\pi(2)$ are estimated by the proportion of couples of primarily alternating orders involving orders at distance 1 and 2, respectively, in the dataset. 
This simple and homogeneous procedure was adopted due to the large number of parameters of some of the models, which can turn the maximization of ${\cal L}$ a difficult problem. In case of Models 1 and 2, that are critical for the conclusions of the article, we used a brute force exploration of the parameter space to make sure that the estimations are accurate enough. The exploration of Model 1 with a resolution of $10^{-4}$ and the exploration of Model 2 with a resolution of $0.005$ were unable to increase ${\cal L}$ and thus supported our previous conclusions. 

}{
}

\end{document}